# From Constraints to Resolution Rules
# Part I : conceptual framework

Denis Berthier
Institut Telecom ; Telecom & Management SudParis
9 rue Charles Fourier, 91011 Evry Cedex, France

**Abstract: Many real world problems appear naturally as constraints satisfaction problems (CSP), for which very efficient algorithms are known. Most of these involve the combination of two techniques: some direct propagation of constraints between variables (with the goal of reducing their sets of possible values) and some kind of structured search (depth-first, breadth-first,…). But when such blind search is not possible or not allowed or when one wants a "constructive" or a "pattern-based" solution, one must devise more complex propagation rules instead. In this case, one can introduce the notion of a candidate (a "still possible" value for a variable). Here, we give this intuitive notion a well defined logical status, from which we can define the concepts of a resolution rule and a resolution theory. In order to keep our analysis as concrete as possible, we illustrate each definition with the well known Sudoku example. Part I proposes a general conceptual framework based on first order logic; with the introduction of chains and braids, Part II will give much deeper results.**

## I. INTRODUCTION

Many real world problems, such as resource allocation, temporal reasoning or scheduling, naturally appear as constraint satisfaction problems (CSP) [1, 2]. Such problems constitute a main sub-area of Artificial Intelligence (AI). A CSP is defined by a finite number of variables with values in some fixed domains and a finite set of constraints (i.e. of relations they must satisfy); it consists of finding a value for each of these variables, such that they globally satisfy all the constraints.

A CSP states the constraints a solution must satisfy, i.e. it says *what* is desired. It does not say anything about *how* a solution can be obtained. But very efficient general purpose algorithms are known [1], which guarantee that they will find a solution if any. Most of these algorithms involve the combination of two very different techniques: some direct propagation of constraints between variables (in order to reduce their sets of possible values) and some kind of structured search with "backtracking" (depth-first, breadth-first,…), consisting of trying (recursively if necessary) a value for a variable, propagating the consequences of this tentative choice and eventually reaching a solution or a contradiction allowing to conclude that this value is impossible.

But, in some cases, such blind search is not possible (for practical reasons, e.g. one wants to simulate human behaviour or one is not in a simulator but in real life) or not allowed (for theoretical or æsthetic reasons, or because one wants to understand what happens, as is the case with most Sudoku players) or one wants a "constructive" solution.

In such situations, it is convenient to introduce the notion of a candidate, i.e. of a "still possible" value for a variable. But a clear definition and a logical status must first be given to this intuitive notion. When this is done, one can define the concepts of a *resolution rule* (a logical formula in the "condition => action" form, which says what to do in some observable situation described by the condition pattern), a *resolution theory*, a *resolution strategy*. One can then study the relation between the original CSP problem and various of its resolution theories. One can also introduce several properties a resolution theory can have, such as confluence (in Part II) and completeness (contrary to general purpose algorithms, a resolution theory cannot in general solve all the instances of a given CSP; evaluating its scope is thus a new topic in its own). This "pattern-based" approach was first introduced in [3], in the limited context of Sudoku solving.

Notice that resolution rules are typical of the kind of rules that can be implemented in an inference engine and resolution theories can be seen as "production systems" [4]. See Part II.

In this paper, we deal only with the case of a finite number of variables with ranges in finite domains and with first order constraints (i.e. constraints between the variables, not between subsets of variables).

This paper is self-contained, both for the general concepts and for their illustrations with the Sudoku example, although deeper results specific to the introduction of chains or to this example will appear in Part II. Section II introduces the first order logical formulation of a general CSP. Section III defines the notion of a candidate and analyses its logical status. Section IV can then define resolution rules, resolution paths, resolution theories and the notion of a pure logic constructive solution. Section V explains in what sense a resolution theory can be incomplete even if its rules seem to express all the constraints in the CSP.

II. THE LOGICAL THEORY ASSOCIATED WITH A CSP

Consider a fixed CSP for n variables $x_1, x_2, \ldots, x_n$ in finite domains $X_1, X_2, \ldots, X_n$, with first order constraints. The CSP can obviously be written as a First Order Logic (FOL) theory (i.e. as a set of FOL axioms expressing the constraints) [1]. Thanks to the equivalence between FOL and Multi-Sorted First Order Logic (MS-FOL) [5], it can also be written as an MS-FOL theory. CSP solutions are in one-to-one correspondence with MS-FOL models of this theory.

In MS-FOL, for each domain $X_k$, one introduces a sort (i.e. a type) $X_k$ (there can be no confusion in using the same letter for the sort and the domain) and a predicate $value_k(x_k)$, with intended meaning "the value of the k-th variable is $x_k$". All the basic functions and predicates necessary to express the given constraints are defined formally as being sorted, so that one doesn't have to write explicit conditions about the sorts of the variables mentioned in a formulæ. This has many advantages in practice (such as keeping formulæ short). The formulæ of our MS-FOL theory are defined as usual, by induction (combining atomic formulæ built on the above basic predicates and functions with logical connectives: and, or, not, typed quantifiers). We can always suppose that, for each value a variable can have, there is a constant symbol of the appropriate sort to name it. We can also adopt a *unique names assumption* for constant symbols: two different constant symbols of the same sort do not designate the same entity. However, no unique names assumption is made for variables. With the following Sudoku example, details missing in the above two paragraphs will hopefully become clearer than through additional logical formalism.

*A. The Sudoku CSP*

Sudoku is generally presented as follows (Fig. 1): given a 9x9 *grid*, partially filled with *numbers* from 1 to 9 (the "entries" or "clues" or "givens" of the problem), complete it with numbers from 1 to 9 in such a way that in each of the nine *rows*, in each of the nine *columns* and in each of the nine disjoint *blocks* of 3x3 contiguous *cells*, the following property holds: there is at most one occurrence of each of these numbers. Notice that this is a special case of the Latin Squares problem (which has no constraints on blocks).

It is natural to consider the three dimensional space with coordinates (n, r, c) and any of the 2D spaces: rc, rn and cn. Moreover, in rc-space, due to the constraint on blocks, it is convenient to introduce an alternative block-square coordinate system [b, s] and variables $X_{bs}$ such that $X_{rc} = X_{bs}$ whenever (r, c) and [b, s] are the coordinates of the same cell. For symmetry reasons, in addition to these variables with values in Numbers, we define additional $X_{rn}$, $X_{cn}$ and $X_{bn}$ variables, with values, respectively in Rows, Columns and Squares, and such that:

$$X_{rc} = n \Leftrightarrow X_{rn} = c \Leftrightarrow X_{cn} = r \Leftrightarrow X_{bn} = s.$$

|   |   |   |   |   |   |   | 3 | 1 |
|---|---|---|---|---|---|---|---|---|
|   |   |   |   | 7 | 9 |   |   |   |
|   |   |   |   |   |   |   |   |   |
|   | 1 | 3 | 2 |   |   |   |   |   |
|   |   | 4 |   |   |   | 7 |   |   |
|   |   |   | 1 |   |   |   |   |   |
| 5 |   |   |   | 4 |   | 6 | 7 |   |
| 2 | 8 |   |   |   |   |   |   |   |
|   |   |   | 3 |   |   |   |   |   |

Figure 1: A typical Sudoku puzzle

Since rows, columns and blocks play similar roles in the defining constraints, they will naturally appear to do so in many other places and it is convenient to introduce a word that makes no difference between them: a *unit* is either a row or a column or a block. And we say that two rc-cells *share a unit* if they are either in the same row or in the same column or in the same block (where "or" is non exclusive). We also say that these two cells are *linked*. It should be noticed that this (symmetric) relation between two cells does not depend in any way on the content of these cells but only on their place in the grid; it is therefore a straightforward and quasi physical notion.

Formulating the Sudoku CSP as an MS-FOL theory is done in three stages: Grid Theory, General Sudoku Theory, Specific Sudoku Puzzle Theory. For definiteness, we consider standard Sudoku only, on a 9x9 grid.

Most CSP problems can similarly be decomposed into three components: axioms for a general and static context (here, the grid) valid for all the instances of the problem, axioms for the general CSP constraints expressed in this context (here, the Sudoku constraints) and axioms for specific instances of the problem (here, the entries of a puzzle).

*B. Grid Theory*
*B.1 The sorts in Grid Theory*

The characteristic of MS-FOL is that it assumes the world of interest is composed of different types of objects, called sorts. In the very limited world of Grid Theory (GT) and of Sudoku Theory (ST), we need only five sorts: Number, Row, Column, Block, Square. Row, Column and Block correspond in the obvious way to rows, columns and blocks, whereas Square corresponds to the relative position of a cell in a block.

Attached to each sort, there are two sets of symbols, one for naming constant objects of this sort, and one for naming variables of this sort. In the GT case, the variables for Numbers are n, n', n'', $n_0$, $n_1$, $n_2$, …; the constants for Numbers are $1_n, 2_n, 3_n, 4_n, 5_n, 6_n, 7_n, 8_n, 9_n$. The variables for Rows are , r', r'', $r_0$, $r_1$, $r_2$, ……; the constants for Rows are $1_r, 2_r, 3_r, 4_r, 5_r, 6_r, 7_r, 8_r, 9_r$. And similarly for the other sorts. For each of the first five sorts, obvious axioms can express the range of the variable of this sort and the unique names assumption.

In conformance with the MS-FOL conventions, a quantifier such as "∀r" (resp. "∀c", "∀n", …) will always mean "for any row r" (resp. "for any column c", "for any number n", …)

*B.2 Function and predicate symbols of Grid Theory*

Grid Theory has no function symbol. In addition to the five equality predicate symbols ($=_n$, $=_r$,… one for each sort), it has only one predicate symbol: *correspondence*, with arity 4 and signature (Row, Column, Block, Square), with intended meaning for atomic formulæ "correspondence(r, c, b, s)" the natural one. Given these basic predicates, one can define auxiliary predicates, considered as shorthands for longer logical formulæ:

– *same-row*, with signature (Row, Column, Row, Column); "same-row($r_1$, $c_1$, $r_2$, $c_2$)" is defined as a shorthand for: $r_1 =_r r_2$;

– and similarly for *same-column* and *same-block*;

– *same-cell*, with signature (Row, Column, Row, Column); "same-cell($r_1$, $c_1$, $r_2$, $c_2$)" is defined as a shorthand for: $r_1 =_r r_2$ & $c_1 =_c c_2$;

As they have been defined, the auxiliary predicates same-row, same-column, same-block and same-cell all have the same arity and signature: informally, they all apply to couples of cells with row-column coordinates. This is very important because it allows to define an auxiliary predicate with the same arity and signature, applying to couples of cells with row-column coordinates, independent of the type of unit they share (we shall see that, most of the time, this type is irrelevant):

– *share-a-unit*, with signature (Row, Column, Row, Column); "share-a-unit($r_1$, $c_1$, $r_2$, $c_2$)" is defined as a shorthand for: ¬same-cell($r_1$, $c_1$, $r_2$, $c_2$) & [ same-row($r_1$, $c_1$, $r_2$, $c_2$) or same-column($r_1$, $c_1$, $r_2$, $c_2$) or same-block($r_1$, $c_1$, $r_2$, $c_2$)].

Of course, the intended meaning of this predicate is that suggested by its name: the two cells share either a row or a column or a block; notice that a cell is not considered as sharing a unit with itself.

*B.3 Axioms of Grid Theory*

In addition to the 5x36 sort axioms, Grid Theory has 81 axioms expressing the (r, c) to [b, s] correspondence of coordinate systems, such as "correspondence($1_r$, $1_c$, $1_b$, $1_s$)".

As an exercise, one can check that this is enough to define the grid (modulo renamings of rows, columns, …). One can also check that the following formula expresses that row r intersects block b: ∃c∃s correspondence(r, c, b, s).

*C. General Sudoku Theory*

General Sudoku Theory (ST) is defined as an extension of Grid Theory. It has the same sorts as GT. In addition to the predicates of GT, it has the following one: *value*, with signature (Number, Row, Column); the intended meaning of atomic formula "value(n, r, c)" is that number n is the value of cell (r, c), i.e. indifferently: $X_{rc}$ = n, $X_{rn}$ = c, $X_{cn}$ = r or $X_{bn}$ = s. It is convenient to introduce an auxiliary predicate value[], written with square braces, with signature (Number, Block, Square), with the same meaning as value, but in [b, s] instead of (r, c) coordinates; "value[n, b, s]" is defined as a shorthand for:
∃r∃c [correspondence(r, c, b, s) & value(n, r, c)]

*C.1 Axioms of Sudoku Theory*

ST contains the axioms of GT, plus the following, written in a symmetrical form that will be useful in the sequel.

$ST_{rc}$: ∀r∀c∀$n_1$∀$n_2$ {value($n_1$, r, c) & value($n_2$, r, c) ⇒ $n_1$ = $n_2$}
$ST_{rn}$: ∀r∀n∀$c_1$∀$c_2$ {value(n, r, $c_1$) & value(n, r, $c_2$) ⇒ $c_1$ = $c_2$}
$ST_{cn}$: ∀c∀n∀$r_1$∀$r_2$ {value(n, $r_1$, c) & value(n, $r_2$, c) ⇒ $r_1$ = $r_2$}
$ST_{bn}$: ∀b∀n∀$s_1$∀$s_2$ {value[n, b, $s_1$] & value[n, b, $s_2$] ⇒ $s_1$ = $s_2$}
$EV_{rc}$: ∀r∀c∃n value(n, r, c)
$EV_{rn}$: ∀r∀n∃c value(n, r, c)
$EV_{cn}$: ∀c∀n∃r value(n, r, c)
$EV_{bn}$: ∀b∀n∃s value[n, b, s].

The formal symmetries inside each of these two groups of four axioms must be noticed. $ST_{rc}$ expresses that an rc-cell can have only one value (this is never stated explicitly, but this should not be forgotten). $ST_{rn}$ (resp. $ST_{cn}$, $ST_{bn}$) expresses that a value can appear only once in a row (resp. a column, a block); these are the standard constraints. Axiom $EV_{rc}$ (resp. $EV_{rn}$, $EV_{cn}$ and $EV_{bn}$) expresses that, in a solution, every rc- (resp. rn-, cn and bn-) cell must have a value; these conditions generally remain implicit in the usual formulation of Sudoku.

*D. Specific Sudoku Puzzle Theory*

In order to be consistent with various sets of entries, ST includes no axioms on specific values. With any specific puzzle P we can associate the axiom $E_P$ defined as the finite conjunction of the set of all the ground atomic formulæ "value($n_k$, $r_i$, $c_j$)" such that there is an entry of P asserting that number $n_k$ must occupy cell ($r_i$, $c_j$). Then, when added to the axioms of ST, axiom $E_P$ defines the MS-FOL theory of the specific puzzle P.

From the point of view of first order logic, everything is said. A solution of puzzle P (if any) is a model (if any) of theory ST + $E_P$. The only problem is that nothing yet is said about how a solution can be found. This is the reason for introducing candidates and resolution rules.

III. CANDIDATES AND THEIR LOGICAL STATUS

*A. Candidates*

If one considers the way Sudoku players solve puzzles, it appears that most of them introduce candidates in the form of "pencil marks" in the cells of the grid. Intuitively, a candidate is a "still possible" value for a cell; candidates in each cell are progressively eliminated during the resolution process.

This very general notion can be introduced for any CSP problem. Unfortunately, it has no *a priori* meaning from the MS-FOL point of view. The reason is not the non-monotonicity of candidates, i.e. that they are progressively withdrawn whereas one can only add information by applying the axioms of a FOL theory: this could easily be dealt with by introducing non-candidates (or impossible values) instead. The real reason is that the intuitive notion of a candidate (as a "still possible" value) and the way it is used in practice suppose a logic in which this idea of "still possible" is formalised. Different "states of knowledge" must then be considered – and this is typically the domain of epistemic logic. We shall therefore adopt *a priori* the following framework, supporting a natural epistemic interpretation of a candidate.

For each variable $x_k$ of the CSP, let us introduce a predicate $cand_k(x_k)$ with intended meaning "the value $x_k$ from domain $X_k$ is not yet known to be impossible for the k-th variable".

The interesting point is that we shall be able to come back to ordinary (though constructivist or intuitionistic) logic for candidates (and thus forget the complexities of epistemic logic).

*B. Knowledge states and knowledge space*

Given a fixed CSP, define a *knowledge state* as any set of values and candidates (formally written as $value_k$ and $cand_k$ predicates). A knowledge state is intended to represent the totality of the ground atomic facts (in terms of values and candidates) that are present in *some* possible state of reasoning for some instance of the CSP. (Invariant background knowledge, such as grid facts in Sudoku, is not explicitly included).

It should be underlined that this notion of a knowledge state has a very concrete and intuitive meaning: for instance, in Sudoku, it represents the situation on a grid with candidates at some point in some resolution process for some puzzle; this is usually named the PM, the "Pencil Marks". (Notice that some knowledge states may be contradictory – so that inconsistent sets of entries can be dealt with).

Let **KS** be the (possibly large, but always finite) set of all possible knowledge states. On **KS**, we define the following order relation: $KS_1 \leq KS_2$ if and only if, for any constant x° (of sort X) one has:

- if $value_X(x°)$ is in $KS_1$, then $value_X(x°)$ is in $KS_2$,
- if $cand_X(x°)$ is in $KS_2$, then $cand_X(x°)$ is in $KS_1$.

If "$KS_1 \leq KS_2$" is intuitively interpreted as "$KS_2$ may appear after $KS_1$ in some resolution process", these conditions express the very intuitive idea that values can only be added and candidates can only be deleted during a resolution process.

For any instance P of the CSP (e.g. for any puzzle P), one can also define the initial knowledge state $KS_P$ corresponding to the starting point of any resolution process for P. Its values are all the entries of P; its candidates are all the possible values of the remaining variables. The set $\mathbf{KS_P} = \{KS / KS_P \leq KS\}$ is thus the set of knowledge states one can reach when starting from P; we call it the *epistemic model* of P.

*C. Knowledge states and epistemic logic*

The above notion of a knowledge state appears to be a particular case of the general concept of a possible world in modal logic; the order relation on the set of knowledge states corresponds to the accessibility relation between possible worlds and our notion of an epistemic model coincides with that of a Kripke model [6]. Let K be the "epistemic operator", i.e. the formal logical operator corresponding to knowing (for any proposition A, KA denotes the proposition "it is known that A" or "the agent under consideration knows that A"). Then, for any proposition A, we have Hintikka's interpretation of KA [7]: in any possible world compatible with what is known (i.e. accessible from the current one), it is the case that A.

Several axiom systems have appeared for epistemic logic (in increasing order of strength: S4 < S4.2 < S4.3 < S4.4 < S5). Moreover, it is known that there is a correspondence between the axioms on the epistemic operator K and the properties of the accessibility relation between possible worlds (this is a form of the classical relationship between syntax and semantics). As the weakest S4 logic is enough for our purposes, we won't get involved in the debates about the best axiomatisation. S4 formalises the following three axioms:

– $KA \Rightarrow A$: "if a proposition is known then it is true" or "only true propositions can be known"; it means that we are speaking of knowledge and not of belief and this supposes the agent (our CSP solver) does not make false inferences; this axiom corresponds to the accessibility relation being reflexive (for all KS in **KS**, one has: $KS \leq KS$);

– $KA \Rightarrow KKA$: (reflection) if a proposition is known then it is known to be known (one is aware of what one knows); this axiom corresponds to the accessibility relation being transitive (for all $KS_1$, $KS_2$ and $KS_3$ in **KS**, one has: if $KS_1 \leq KS_2$ and $KS_2 \leq KS_3$, then $KS_1 \leq KS_3$);

– $K(A \Rightarrow B) \Rightarrow (KA \Rightarrow KB)$: (limited deductive closure of knowledge) if it is known that $A \Rightarrow B$, then if it is known that A, then it is known that B. In the case of CSP, this will be applied as follows: when a resolution rule [$A \Rightarrow B$] is known [$K(A \Rightarrow B)$], if its conditions [A] are known to be satisfied [KA] then its conclusions [B] are known to be satisfied [KB].

*D. Values and candidates*

As we want our resolution rules to deal with candidates, all our initial MS-FOL concepts must be re-interpreted in the context of epistemic logic.

The entries of the problem P are not only true in the initial knowledge state $KS_P$, they are known to be true in this state:

they must be written as $Kvalue_k$; similarly, the initial candidates for a variable are not only the *a priori* possible values for it; they must be interpreted as not yet known to be impossible: $\neg K \neg cand$.

Moreover, as a resolution rule must be effective, it must satisfy the following: a condition on the absence of a candidate must mean that it is effectively known to be impossible: $K\neg cand$; a condition on the presence of a candidate must mean that it is not effectively known to be impossible: $\neg K\neg cand$; a conclusion on the assertion of a value must mean that this value becomes effectively known to be true: $Kvalue$; a conclusion on the negation of a candidate must mean that this candidate becomes effectively known to be impossible: $K\neg cand$.

As a result, in a resolution rule, a predicate "$value_k$" will never appear alone but only in the construct "$Kvalue_k(x_k)$"; a predicate "$cand_k$" will never appear alone but only in the construct $\neg K\neg cand_k$ (given that $K\neg cand_k$ is equivalent, in any modal theory, to $\neg\neg K\neg cand_k$).

All this entails that we can use well known correspondences of modal logic S4 with intuitionistic logic [9] and constructive logic [10] to "forget" the K operator (thus merely replacing everywhere "$Kvalue_k$" with "$value_k$" and "$\neg K\neg cand_k$" with "$cand_k$"), provided that we consider that we are now using intuitionistic or constructive logic. We have thus eliminated the epistemic operator that first appeared necessary to give the notion of a candidate a well defined logical status. Said otherwise: *at the very moderate price of using intuitionnistic or constructive logic, in spite of the fact that candidates can be given an epistemic status, no explicit epistemic operator will ever be needed in the logical formulation of resolution rules*.

One thing remains to be clarified: the relation between values and candidates. In the epistemic interpretation, a value $a_k$ for a variable $x_k$ is known to be true if and only if all the other possible values for this variable are known to be false:
$\forall x_k [Kvalue_k(x_k) \Leftrightarrow \forall x'_k \neq x_k\ K\neg cand_k(x'_k)]$.

Using the equivalence between $K\neg$ and $\neg\neg K\neg$ and forgetting the K operator as explained above, we get the value-to-candidate-relation intuitionistic axiom, for each variable $x_k$:
**$VCR_k$: $\forall x_k [value_k(x_k) \Leftrightarrow \forall x'_k \neq x_k \neg cand_k(x'_k)]$.**

IV. RESOLUTION RULES AND RESOLUTION THEORIES

*A. General definitions*

Definiton: a formula in the MS-FOL language of a CSP is in *the condition-action form* if it is written as $A \Rightarrow B$, possibly surrounded with quantifiers, where A does not contain explicitly the "$\Rightarrow$" sign and B is a conjunction of value predicates and of negated cand predicates (no disjunction is allowed in B); all the variables appearing in B must already appear in A and be universally quantified.

Definitons: a formula in the condition-action form is a *resolution rule* for a CSP if it is an intuitionistically (or constructively) valid consequence of the CSP axioms and of VCR. A *resolution theory* for a CSP is a set of resolution rules. Given a resolution theory T, a *resolution path* in T for an instance P of the CSP is a sequence of knowledge states starting with P and such that each step is justified by a rule in T. A *resolution theory T solves an instance P of the CSP* if one can exhibit a resolution path in T leading to a solution. Notice that, contrary to the general notion of a solution of a CSP as any model of the associated FOL or MS-FOL theory, this is a restrictive definition of a solution; it can be called a *constructive definition* of a solution: a resolution theory solves an instance of the CSP only if it does so in the constructive way defined above.

*B. The Basic Resolution Theory of a CSP*

For any CSP, there is a Universal Resolution Theory, URT, defined as the union of the following three types of rules, which are the mere re-writing of each of the $VCR_k$ axioms, for each sort:

– Elementary Constraints Propagation rule for sort $X_k$:
$ECP_k$: $\forall x_k \forall x_{k1} \neq x_k \{value_k(x_k) \Rightarrow \neg cand_k(x_{k1})\}$;

– "Singles" rule for sort $X_k$:
$S_k$: $\forall x_k\{[cand_k(x_{k1})\ \&\ \forall x_{k1} \neq x_k \neg cand_k(x_{k1})] \Rightarrow value_k(x_k)\}$.

– Contradiction Detection for sort $X_k$:
$CD_k$: $\forall x_k(\neg value(x_k)\ \&\ \neg cand_k(x_k)) \Rightarrow \bot$, where "$\bot$" is any false formula.

But for any CSP, there is also a Basic Resolution Theory, BRT, which is the union of URT with all the rules specific to this CSP expressing the direct contradictions (if any) between its different variables (see Part II). All the resolution theories we shall consider will be extensions of this BRT.

*C. Example from Sudoku*

The BRT of the Sudoku CSP (say BSRT) consists of the following rules. The elementary constraints propagation (ECP) rules are the re-writing of the left-to-right part of axioms $ST_{rc}$, $ST_{bn}$, $ST_{cn}$ and $ST_{bn}$ in the condition-action form:
ECP1: $\forall r \forall c \forall n \forall n_1 \neq n\ \{value(n, r, c) \Rightarrow \neg cand(n_1, r, c)\}$
ECP2: $\forall r \forall n \forall c \forall c_1 \neq c\ \{value(n, r, c) \Rightarrow \neg cand(n, r, c_1)\}$
ECP3: $\forall c \forall n \forall r \forall r_1 \neq r\ \{value(n, r, c) \Rightarrow \neg cand(n, r_1, c)\}$
ECP4: $\forall b \forall n \forall s \forall s_1 \neq s\ \{value[n, b, s] \Rightarrow \neg cand[n, b, s_1]\}$

Here cand[] is related to cand() in the same way as value[] was related to value().

The well-known rules for Singles ("Naked-Singles" and "Hidden-Singles") are the re-writing of the right-to-left part of the same axioms:
NS: $\forall r \forall c \forall n\{\ [cand(n, r, c)\ \&\ \forall n_1 \neq n\ \neg cand(n_1, r, c)]$
$\Rightarrow value(n, r, c)\}$

$HS_{rn}$: $\forall r \forall n \forall c\{$ [cand(n, r, c) & $\forall c_1 \neq c$ ¬cand(n, r, $c_1$)] $\Rightarrow$ value(n, r, c)$\}$

$HS_{cn}$: $\forall c \forall n \forall r\{$ [cand(n, r, c) & $\forall r_1 \neq r$ ¬cand(n, $r_1$, c)] $\Rightarrow$ value(n, r, c)$\}$

$HS_{bn}$: $\forall b \forall n \forall s\{$ [cand'[n, b, s] & $\forall s_1 \neq s$ ¬ cand'[n, b, $s_1$] ] $\Rightarrow$ value'[n, b, s]$\}$

Axioms EV have a special translation, with meaning: if there is a (rc-, rn- cn- or bn-) cell for which no value remains possible, then the problem has no solution:

$CD_{rc}$: $\exists r \exists c \forall n$[¬value(n, r, c) & ¬cand(n, r, c)] $\Rightarrow \bot$.
$CD_{rn}$: $\exists r \exists n \forall c$[¬value(n, r, c) & ¬cand(n, r, c)] $\Rightarrow \bot$.
$CD_{cn}$: $\exists c \exists n \forall r$[¬value(n, r, c) & ¬cand(n, r, c)] $\Rightarrow \bot$.
$CD_{bn}$: $\exists b \exists n \forall s$[¬value'(n, b, s) & ¬cand'(n, b, s)] $\Rightarrow \bot$.

## V. COMPLETENESS

### A. Completeness

Now that the main concepts are defined, we can ask: what does it mean for a Resolution Theory T for a given CSP to be "complete"? Notice that all the results that can be produced (i.e. all the values that can be asserted and all the candidates that can be eliminated) when a resolution theory T is applied to a given instance P of the CSP are logical consequences of theory T $\cup$ $E_P$ (where $E_P$ is the conjunction of the entries for P); these results must be valid for any solution for P (i.e. for any model of T $\cup$ $E_P$). Therefore a resolution theory can only solve instances of the CSP that have a unique solution and one can give three sensible definitions of the completeness of T: 1) it solves all the instances that have a unique solution; 2) for any instance, it finds all the values common to all its solutions; 3) for any instance, it finds all the values common to all its solutions and it eliminates all the candidates that are excluded by any solution.

Obviously, the third definition implies the second, which implies the first, but whether the converse of any of these two implications is true in general remains an open question.

### B. Why a Basic Resolution Theory may not be enough

In the case of Sudoku, one may think that the obvious resolution rules of BSRT are enough to solve any puzzle. After all, don't they express all that there is in the axioms? It is important to understand in what sense they are not enough and why.

*These rules are not enough because our notion of a solution within a resolution theory T a priori restricts them to being used constructively*; said otherwise, we look only for models of T obtained constructively from the rules in T; but a solution of a CSP is any model of the original axioms, whether it is obtained in a constructive way or not (it may need some "guessing").

To evaluate how far these rules are from being enough, they have been implemented in an inference engine (CLIPS) and a statistical analysis has been made on tens of thousands of randomly generated puzzles. It shows that they can solve 42% of the minimal puzzles ("minimal" means "has a unique solution and has several solutions if any entry is deleted"; statistics would be meaningless without this condition).

## VI. CONCLUSION

We have proposed a general conceptual framework for approximating a constraint satisfaction problem with constructive resolution theories in which the intuitive notion of a candidate is given a simple and well defined logical status with an underlying epistemic meaning. We have given a detailed illustration of these concepts with the Sudoku CSP. We have explained why a resolution theory, even though it seems to express all the constraints of the CSP, may not be complete.

One may ask: is using a resolution theory more efficient (from a computational point of view) than combining elementary constraints propagation with blind search? In the Sudoku example, our simulations (see Part II) show that the answer is clearly negative; moreover, as there exist very efficient general purpose search algorithms, we think this answer is general. But, instead of setting the focus on computational efficiency, as is generally the case, our approach sets the focus on constructiveness of the solution. Another general question remains open: how "close" can one approximate a CSP with a well chosen resolution theory? We have no general answer. But, in Part II of this paper, we shall define elaborated resolution theories, give a meaning to the word "close" and provide detailed results for the Sudoku CSP.